%% file: main.tex
\documentclass[10pt,twocolumn,letterpaper]{article}
\usepackage[accsupp]{axessibility}  % Improves PDF readability for those with disabilities.
\usepackage{float}
\usepackage{multirow}
\usepackage{makecell}
\usepackage{booktabs}
\usepackage{colortbl}
\usepackage{cvpr}              % To produce the CAMERA-READY version

\definecolor{cvprblue}{rgb}{0.21,0.49,0.74}
\usepackage[pagebackref,breaklinks,colorlinks,allcolors=cvprblue]{hyperref}
\usepackage{booktabs}
\def\paperID{6483} % *** Enter the Paper ID here
\def\confName{CVPR}
\def\confYear{2025}

\title{Monocular and Generalizable Gaussian Talking Head Animation}

\author{
Shengjie Gong$^{1}$\ \ \
Haojie Li$^1$\ \ \
Jiapeng Tang$^2$\ \ \
Dongming Hu$^{1}$\ \ \
Shuangping Huang$^{1,3}$\footnotemark[2]\\
Hao Chen$^1$\ \ \
Tianshui Chen$^4$\ \ \
Zhuoman Liu$^5$
 \\ {
$^1$South China University of Technology~~
$^2$Technical University of Munich ~~  $^3$Pazhou Laboratory} \\
{
$^4$Guangdong University of Technology~~~~~~~~
$^5$The Hong Kong Polytechnic University}
}

\begin{document}
\renewcommand{\thefootnote}{\fnsymbol{footnote}} %将脚注符号设置为fnsymbol类型，即特殊符号表示
% \maketitle
\twocolumn[{
\maketitle
\begin{center}
    \captionsetup{type=figure}
    \vspace{-2em}
    \includegraphics[width=1.\textwidth]{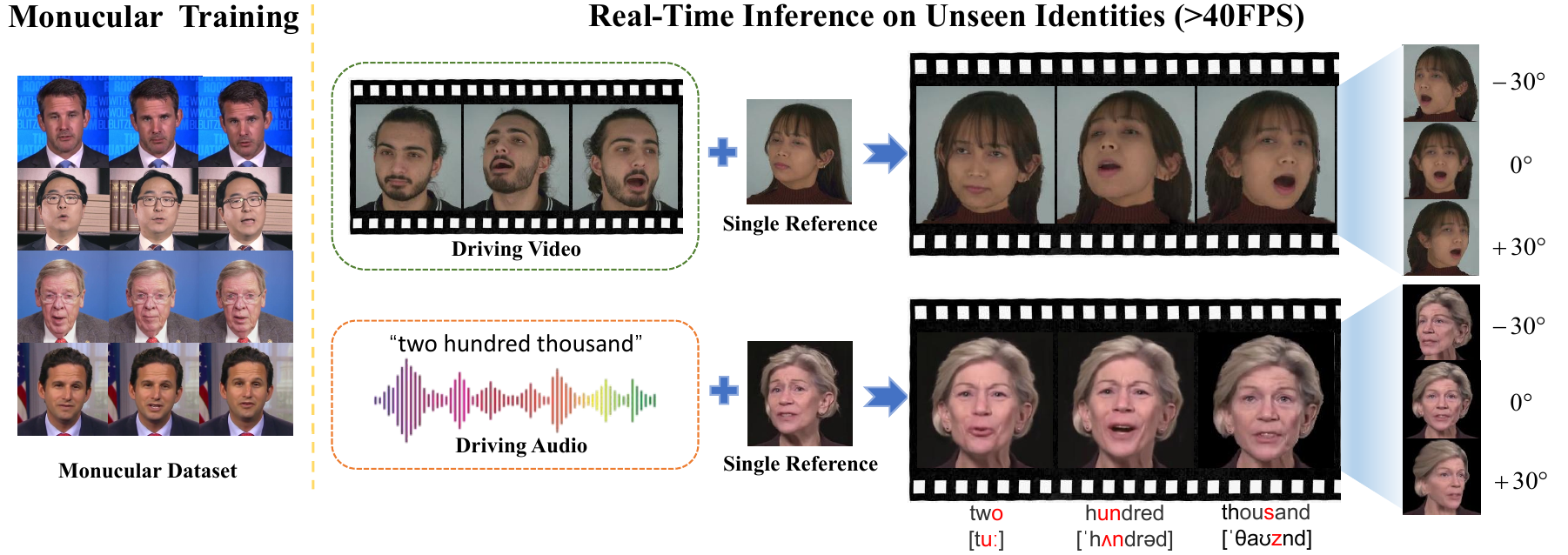}
    \vspace{-2em}
    \captionof{figure}{
    MGGTalk is trained using only monocular datasets and enable generalizing to unseen identities without personalized re-training. Additionally, it supports real-time generation of talking heads in diverse poses and novel viewpoints.
    } 
\end{center}
}]

\footnotetext[2]{Corresponding author.} %对应脚注[2]
\input{sec/0_abstract} 
\input{sec/1_intro}
\input{sec/2_related}
\input{sec/3_method}
\input{sec/4_exp}
\input{sec/5_conclusion}
\clearpage
\renewcommand{\thesection}{\Alph{section}}
\setcounter{section}{0}         % 重置计数器
\input{sec/X_suppl}

\clearpage
\bibliographystyle{unsrt}
\setcitestyle{numbers}
\bibliography{main}

\end{document}

%% file: sec/0_abstract.tex
\begin{abstract}
In this work, we introduce Monocular and Generalizable Gaussian Talking Head Animation (MGGTalk), which requires monocular datasets and generalizes to unseen identities without personalized re-training.
Compared with previous 3D Gaussian Splatting (3DGS) methods that requires elusive multi-view datasets or tedious personalized learning/inference, MGGtalk enables more practical and broader applications.
However, in the absence of multi-view and personalized training data, the incompleteness of geometric and appearance information poses a significant challenge. 
To address these challenges, MGGTalk explores depth information to enhance geometric and facial symmetry characteristics to supplement both geometric and appearance features. 
Initially, based on the pixel-wise geometric information obtained from depth estimation, we incorporate symmetry operations and point cloud filtering techniques to ensure a complete and precise position parameter for 3DGS.
Subsequently, we adopt a two-stage strategy with symmetric priors for predicting the remaining 3DGS parameters. We begin by predicting Gaussian parameters for the visible facial regions of the source image. These parameters are subsequently utilized to improve the prediction of Gaussian parameters for the non-visible regions.
Extensive experiments demonstrate that MGGTalk surpasses previous state-of-the-art methods, achieving superior performance across various metrics.
Project page: {\url{https://scut-mmpr.github.io/MGGTalk-Homepage/}}.
\end{abstract}
\vspace{-5mm}

%% file: sec/1_intro.tex
\section{Introduction}
\label{sec:intro}
One-shot talking head generation synthesizes realistic talking head videos from a single reference image and driving sources like audio or motion. This technology is promising for applications such as video dubbing \cite{kr2019towards, prajwal2020lip, xie2021towards}, film production \cite{zhou2020makelttalk}, and video conferencing \cite{chen2020talking, wang2021one}.
A key challenge is generating high-quality, motion-rich videos from a single image, driven by expressive and pose-related information while ensuring robust identity protection.

Existing methods for talking head animation can be broadly categorized into 2D generator \cite{zhang2023sadtalker, liu2024anitalker, chen2024echomimic, prajwal2020lip, yin2022styleheat, hong2022depth,xu2024self,chen2025contrastive} and 3D rendering \cite{deng2024portrait4d, shao2024splattingavatar, qian2024gaussianavatars,zheng2024headgap, ye2024real3d, li2024generalizable,chen2024learning} approaches. 
2D generator methods typically use generative network, they often lack 3D modeling of the face and attribute disentanglement, which affects the overall generation quality.
Some 3D rendering methods \cite{ye2024real3d, li2024generalizable, deng2024portrait4d} utilize Neural Radiance Fields (NeRF) to synthesize more lifelike talking head videos by using 3D facial modeling. 
However, these NeRF-based approaches often suffer from high computational complexity, making real-time rendering challenging.
In this context, 3D Gaussian Splatting (3DGS) \cite{kerbl20233d} emerges as a promising solution, featuring a flexible, explicit, and high-fidelity 3D representation that allows for high-quality and fast rendering.

\begin{figure}[t]
\hspace{-3mm}
%\centering
\includegraphics[width=0.5\textwidth]{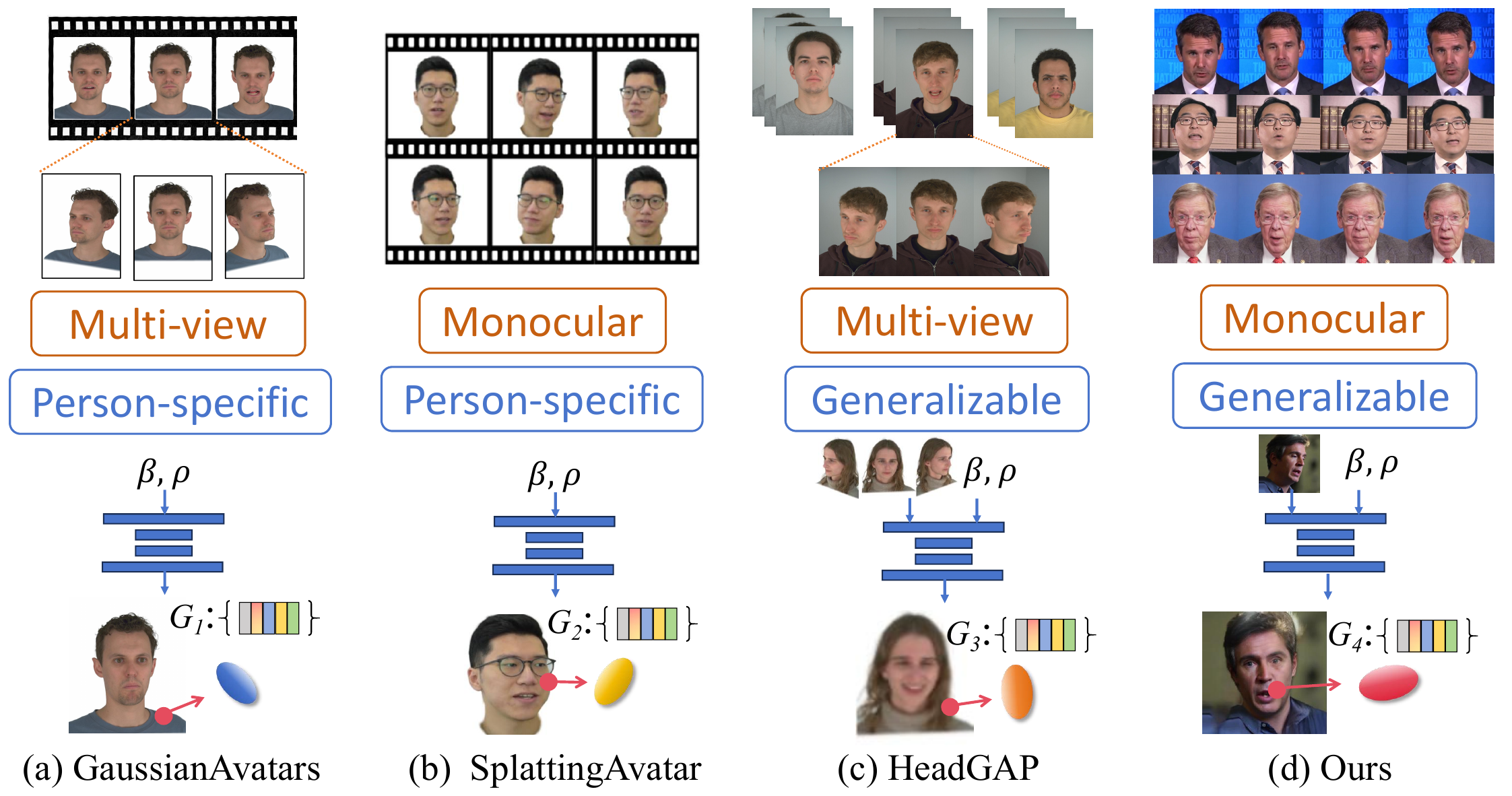}
\vspace{-7mm}
\caption{Comparison of 3DGS-based talking head animation methods, which typically rely on elusive multi-view datasets or tedious personalized learning.
Our method achieves generalization to unseen identities while only training on monocular datasets.
}
\vspace{-5mm}
\label{fig:dataset}
\end{figure}

However, as shown in Figure \ref{fig:dataset}, many current 3DGS-based methods \cite{xu2024gaussian, shao2024splattingavatar, qian2024gaussianavatars,teotia2024gaussianheads,yu2024gaussiantalker} rely on multi-view datasets or personalized training to achieve high-quality results, which complicates generalization.
Several methods \cite{qian2024gaussianavatars, teotia2024gaussianheads} generate realistic outputs but still depend on these multi-view datasets and personalized training approaches.
To reduce data requirements, some approaches \cite{xu2024gaussian, shao2024splattingavatar, yu2024gaussiantalker} utilize only monocular video data for personalized generation.
Despite advancing data efficiency and accessibility, these methods still struggle to generalize to unseen individuals without additional training or fine-tuning.
Another method \cite{zheng2024headgap} aims to balance data efficiency and generalization through few-shot learning techniques, generating satisfactory results with limited input. 
However, they still require multi-view data during training to build robust models, which constrains their effectiveness in real-world applications.

We propose the Monocular and Generalizable Gaussian Talking Head Animation (MGGTalk) framework, which is trained on monocular datasets and generalizes to new identities without additional fine-tuning.
We exploit facial characteristics, focusing on depth estimation for capturing detailed facial geometry and leveraging prior knowledge of facial symmetry to improve the completeness of both geometric structure and texture.
By integrating these characteristics, we address the incomplete feature representation often caused by monocular data limitations, generating more complete 3D Gaussian facial attributes similar to those achieved through multi-view or personalized learning approaches. 
In this way, we achieve a generalizable gaussian for high-quality, multi-view consistent talking head animation using only monocular data.
Specifically, we introduce two key modules: Depth-Aware Symmetric Geometry Reconstruction (DSGR), Symmetric Gaussian Prediction (SGP)
. In the DSGR module, thanks to the advancements in monocular depth estimation, we use the pixel-wise depth estimation data obtained from Geowizard \cite{fu2025geowizard} as the initial information for 3D positions.
Subsequent this initial capture is followed by a refinement network that mitigates common depth-related discrepancies, enhancing the precision of the resulting point cloud. 
To address the challenge of incomplete geometric information caused by extreme facial poses, we implement a symmetry operation in the canonical pose space to complement the geometry of regions that are invisible in the source image.
Finally, we introduce a voxel-based filter that quantizes point cloud coordinates, effectively reducing redundancy and overlap in the reconstructed point cloud.
Then, the SGP module is designed to better utilize geometric information and texture features to predict additional Gaussian parameters.
Initially, we predict the remaining Gaussian parameters for the point cloud representing the visible facial regions of the source image, which benefits from labeled supervision, enabling more precise expression learning.
Subsequently, for the non-visible areas derived from the symmetry operation in the DSGR model, we utilize the Gaussian parameters from the visible areas as inputs to guide the prediction process, thereby reducing the difficulty of predicting parameters for these unlabeled regions. 
This approach ensures that we obtain more complete and precise Gaussian parameters for the entire facial model.

Our main contributions are as follows:
\begin{itemize}
    \item We propose the MGGTalk framework, which explores depth and symmetric facial characteristics for effective 3DGS-based talking head animation. It enables training on monocular datasets and generalizes to unseen identities without personalized re-training. To our knowledge, this is the first attempt to achieve this goal.
    \item We design the Depth-Aware Symmetric Geometry Reconstruction and Symmetric Gaussian Prediction modules, which leverage depth and symmetry information to enhance both geometric and appearance features for effective 3D Gaussian reconstruction and high-quality rendering. 
    \item We conduct extensive experiments and ablative studies, demonstrating the effectiveness of MGGTalk in both audio-driven and video-driven talking head generation and exhibiting the actual contribution of each module.
\end{itemize}

%% file: sec/2_related.tex
\section{Related Work}
\label{sec:related}

\subsection{Talking Head Generation with 2D Generator}
Many talking head generation methods utilize generative adversarial networks (GAN) \cite{wang2022one, zhang2023sadtalker, zhong2023identity, yin2022styleheat, ren2021pirenderer, hong2022depth, ma2023styletalk} or diffusion models \cite{shen2023difftalk, ma2023dreamtalk, liu2024anitalker, chen2024echomimic, zeng2023face,bigata2025keyface,dai2023disentangling,dai2024one} as 2D generators to produce high-quality talking head videos.
Some methods represent facial motion using keypoints \cite{zhong2023identity, ren2021pirenderer, hong2022depth}, while others disentangle facial attributes in latent space for input to 2D generators \cite{liu2024anitalker}.
However, the lack of 3D head modeling results in issues such as facial distortions and identity inconsistency.
Although some approaches introduce 3D information, such as 3D Morphable Models (3DMMs)~\cite{paysan20093d,tang2024dphms,li2017learning,giebenhain2023learning}, to represent facial geometry or motion \cite{zhang2023sadtalker, ma2023styletalk, zeng2023face}, the transformation of 3D features through 2D generators often results in a loss of geometric information, making it difficult to achieve consistent multi-view talking head generation.
In contrast, our approach explicitly models the 3D face and utilizes point cloud-based rendering to generate multi-view consistent talking head images.
%physically-based rendering
\subsection{Talking Head Generation with 3D Rendering}
Talking head generation methods that employ 3D rendering typically rely on neural or point-based techniques to achieve high-quality head avatars.
Among these, Neural Radiance Fields (NeRF) have been widely adopted for generating detailed and realistic 3D facial reconstructions \cite{li2024s, chu2024gpavatar, deng2024portrait4d, li2024generalizable, ye2024real3d}.
However, the computational complexity of NeRF leads to costly training and inference.

Recently, 3D Gaussian Splatting (3DGS) has demonstrated state-of-the-art visual quality and real-time rendering capabilities for 3D scene reconstruction tasks \cite{kerbl20233d}.
Many methods have successfully applied 3DGS for talking head generation, achieving impressive performance and rendering speed \cite{xu2024gaussian, shao2024splattingavatar, chen2024gstalker, cho2024gaussiantalker, yu2024gaussiantalker,qian2024gaussianavatars,zheng2024headgap,xu20253d,kirschstein2024gghead}. 
Several of these approaches \cite{ xu2024gaussian, yu2024gaussiantalker, shao2024splattingavatar, tang2025gaf} use several minutes of monocular video from a specific individual for training, achieving high visual quality and real-time rendering. 
For example, GAF~\cite{tang2025gaf} utilized multi-view head diffusion priors to reconstruct Gaussian avatars from monocular videos.
SplattingAvatar \cite{shao2024splattingavatar} binds 3DGS to the 3DMM facial model to represent the head appearance and the facial expression changes are then realized by driving the positional offsets of 3DGS through the motion of the 3DMM.
In contrast, GaussianAvatars \cite{qian2024gaussianavatars} utilizes multi-view video to obtain a more accurate 3DMM and attaches 3DGS to the triangular mesh of the 3DMM, achieving improved 3D consistency.
However, most of these methods require multi-view or monocular video from specific identities for training, limiting their ability to generalize to unseen individuals. 
Some methods have introduced 3D Gaussian parametric priors for pre-training, allowing generalization to new input images~\cite{zheng2024headgap,xu20253d}. 
Nevertheless, they require training on multi-view datasets, the acquisition of which is relatively demanding, thereby limiting their applicability.
In this paper, we propose MGGTalk, which requires only monocular data for training and can generalize to unseen identities without additional retraining.

%% file: sec/3_method.tex
\section{MGGTalk}
\label{sec:method}
\begin{figure*}
\vspace{-3mm}
    \centering
    \includegraphics[width=1\linewidth]{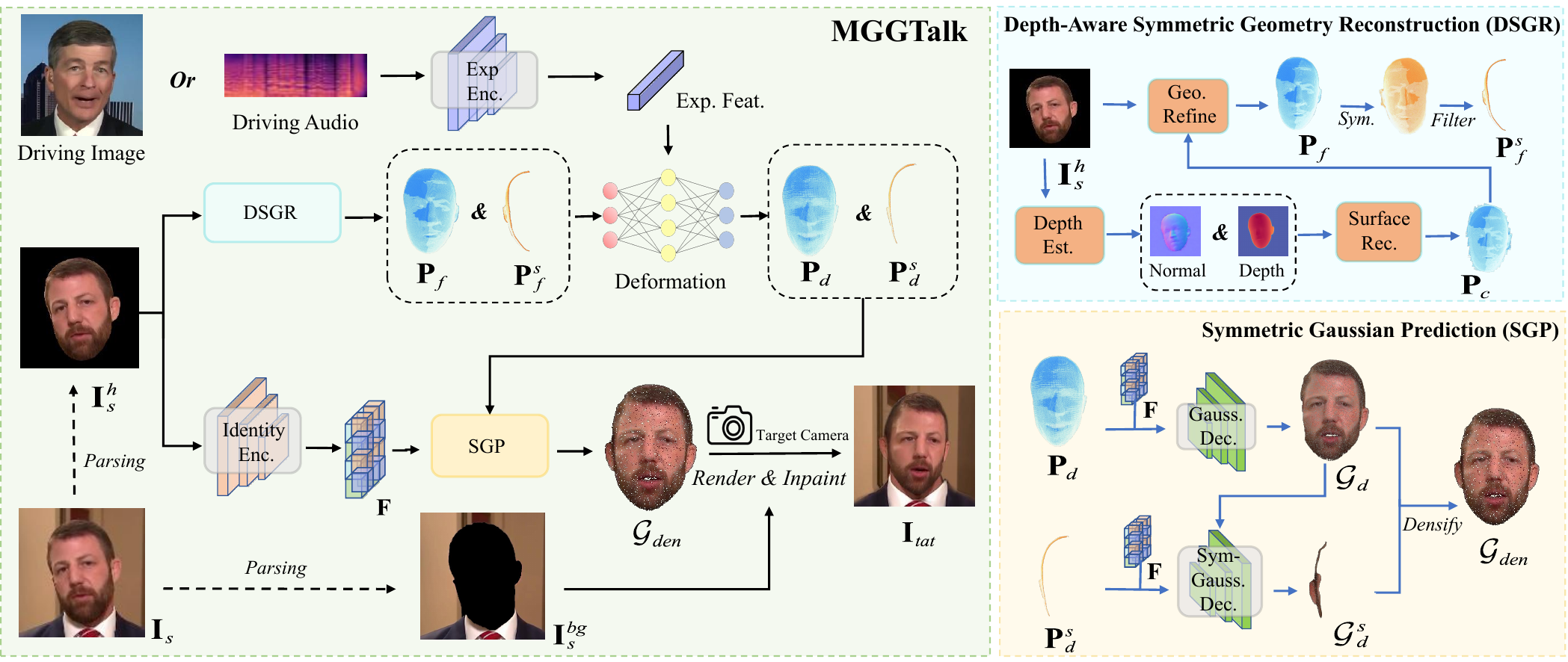}
    \vspace{-7mm}
    \caption{
    \textbf{Pipeline overview of the MGGTalk.} 
Given a source image $\mathbf{I}_s$, we first use semantic parsing to extract the head region $\mathbf{I}_s^{h}$ and torse-background $\mathbf{I}_s^{bg}$. 
    The DGSR module generates point clouds $[\mathbf{P}_f; \mathbf{P}_f^{s}]$ for visible and invisible regions from  $\mathbf{I}_s^{h}$. 
    Expression features from the driving image or audio are used by the Deformation Network to edit the point cloud, resulting in $[\mathbf{P}_{d}; \mathbf{P}_{d}^{s}]$. 
    The SGP module then takes the identity encoding $\mathbf{F}$ from $\mathbf{I}_s^{h}$ and the deformed point cloud $[\mathbf{P}_{d}; \mathbf{P}_{d}^{s}]$ to predict the complete Gaussian parameters $\mathcal{G}_{den}$. 
    Finally,  $\mathcal{G}_{den}$ is  rendered and composited with torso-background $\mathbf{I}_s^{bg}$ to obtain the target $\mathbf{I}_{tat}$.
    }
    
    \label{fig:enter-label}
    \vspace{-5mm}
\end{figure*}

Compared to multi-view or person-specific methods, monocular training with one-shot inference often leads to incomplete Gaussian parameters due to the limited information available. 
To address this challenge, we propose the MGGTalk framework, which uses Depth-Aware Symmetric Geometry Reconstruction (DSGR, Sec. \ref{sec:Depth}) and Symmetric Gaussian Prediction (SGP, Sec. \ref{sec:appear}) modules to leverage facial depth and symmetry priors. 
This allows for the construction of a generalizable 3DGS representation from a single image, enabling high-quality, multi-view consistent one-shot talking head animation.
An overview of our framework is shown in Figure \ref{fig:enter-label}. 
Given a segmented head image from the source, the DSGR module uses depth information and symmetry priors to generate a point cloud representing the visible facial regions and a mirrored point cloud representing the invisible regions. 
Subsequently, an MLP-based deformation network takes expression features obtained from either the driving image or audio encoding as input and adjusts the point clouds, resulting in edited point clouds.
Specifically, expression features are extracted using 3DMM estimation \cite{deng2019accurate} for driving images or an audio-to-expression network \cite{zhang2023sadtalker} for driving audio.
The SGP module then combines the identity encoding and the driven point cloud to generate the complete set of Gaussian parameters.
Finally, the rendering and inpainting process projects the Gaussian parameters onto the target view and completes the torso-background, resulting in the final target image.

\subsection{Depth-Aware Symmetric Geometry  Reconstruction}
\label{sec:Depth}
Reconstructing 3D geometry from a 2D facial image is challenging due to the inherent lack of three-dimensional information, especially when parts of the face are invisible (e.g., in profile views).
To address these challenges, we propose the Depth-Aware Symmetric Geometry Reconstruction module, which incorporates depth information and facial symmetry priors to obtain a more complete facial geometry from a single input image. 
Our approach reconstructs the 3D point clouds for the visible facial regions using depth estimation, while the symmetry operation supplements the geometry of the invisible regions.

For the visible facial regions of the input image, we introduce depth estimation information as a bridge between 2D and 3D.
Some methods \cite{zheng2024gps,chen2024mvsplat} have already utilized depth maps to back-project into 3D space for constructing 3DGS positions. 
However, unlike the above methods that rely on multi-view depth estimation, monocular depth estimation often less accurate and lacks detailed information.
To address this limitation, we employ a surface reconstruction module that takes depth and normal information derived from the depth estimation process to enhance surface geometry and detail capture, providing a more robust geometric representation.
A geometry refinement module is then introduced to achieve a more precise facial geometry.
Specifically, we first derive the depth map and normal map from the input image $\mathbf{I}_s^{h}$ using a pre-trained depth estimation $\Phi_{\text{depth}}$ \cite{fu2025geowizard}. 
We then apply the Bilateral Normal Integration algorithm (BINI) \cite{cao2022bilateral} on the depth map and normal map for surface reconstruction to generate the corresponding coarse facial point cloud $\mathbf{P}_{c}$.
This process can be formalized as follows:
\begin{equation} \label{eq:dep} 
\mathbf{P}_{c} = \mathbf{BINI}(\Phi_{depth}(\mathbf{I}_s^{h})).
\end{equation}
The initial geometric structure is relatively flat and fails to represent the 3D facial structure accurately. 
To address this, we introduce a 2D UNet-based geometry refinement network $\Phi_{refine}$. 
This module takes the coarse geometry $\mathbf{P}_{c}$ and the source image $\mathbf{I}_s^{h}$ as inputs, predicting offset values for the 3D point cloud, which are then added to the coarse geometry to achieve refined corrections $\mathbf{P}_f$: 
\begin{equation} \label{eq:SSG} 
\mathbf{P}_f = \mathbf{P}_{c} + \Phi_{refine}(\mathbf{P}_{c}, \mathbf{I}_s^{h}).
\end{equation}

For the facial regions of the input image that are not visible, we utilize the natural left-right symmetry of the human face for supplementation. 
Specifically, we perform symmetry in the canonical pose space by inverting the x-coordinates of $\mathbf{P}_f$ to obtain a symmetry point cloud.
We then propose a voxel filter $\mathcal{F}_{voxel}$ to remove overlapping regions between the symmetry point cloud and $\mathbf{P}_f$.
The voxel filter quantifies the 3D point cloud coordinates.
Points in the symmetric point cloud $\mathbf{P}_f^{s}$ are removed if they satisfy either of the following conditions: (1) a corresponding point exists within the neighboring voxel of $\mathbf{P}_f$, or (2) they occlude some points in $\mathbf{P}_f$.
The process of obtaining the symmetric point cloud $\mathbf{P}_f^{s}$ can be formulated as follows:
\begin{equation} \label{eq:sym} 
\mathbf{P}_f^{s} = \mathcal{F}_{voxel}( \mathbf{P}_f \cdot \begin{bmatrix}
-1,1,1
\end{bmatrix}^{T})
\end{equation}

\renewcommand{\arraystretch}{1.2}
\begin{table*}[t]
\fontsize{30}{40}\selectfont
\centering
\caption{Quantitative results of video-driven methods on the HDTF dataset \cite{zhang2021flow} and NeRSemble-Mono dataset \cite{kirschstein2023NeRSemble}. 
We use \textbf{bold text} to indicate the best results and \underline{underline} to denote the second-best results.}
\label{tab:video-hdtf}
\vspace{-2mm}
\resizebox{2\columnwidth}{!}{%
\begin{tabular}{cccccccccccccccccccc}
\Xhline{5pt}
\multirow{3}{*}{Method}                                  & \multicolumn{9}{c}{HDTF}                                                                                                                                                                                                                                                 & \multicolumn{1}{l}{} & \multicolumn{9}{c}{NeRSemble-Mono}                                                                                                                            \\ \cline{2-10} \cline{12-20} 
                                                         & \multicolumn{5}{c}{Self-Reenactment}                                                                                                                               &  & \multicolumn{3}{c}{Cross-Reenactment}                                                            &                      & \multicolumn{5}{c}{Self-Reenactment}                                                               &  & \multicolumn{3}{c}{Cross-Reenactment}            \\ \cline{2-6} \cline{8-10} \cline{12-16} \cline{18-20} 
                                                         & PSNR↑                          & SSIM↑                          & FID↓                           & AED↓                           & APD↓                           &  & FID↓                           & AED↓                           & APD↓                           &                      & PSNR↑          & SSIM↑                          & FID↓           & AED↓           & APD↓           &  & FID↓           & AED↓           & APD↓           \\ \cline{1-20} 
Styleheat \cite{yin2022styleheat}       & 30.23                          & 0.664                          & 61.92                          & 0.209                          & 0.245                          &  & 76.43                          & 0.369                          & 0.334                          &                      & 30.67          & 0.719                          & 77.45          & 0.189          & 0.602          &  & 93.25          & 0.345          & 0.854          \\
DaGAN \cite{hong2022depth}              & 30.94                          & 0.729                          & \underline{33.23} & \underline{0.106} & 0.172                          &  & 48.20                          & \underline{0.225} & 0.258                          &                      & 31.39          & 0.745                          & 82.50          & 0.166          & 0.575          &  & 108.12         & 0.316          & 0.770          \\
ROME \cite{khakhulin2022realistic}      & 30.99                          & 0.775                          & 76.44                          & 0.135                          & \underline{0.142} &  & 79.31                         & 0.229                          & 0.286                          &                      & 31.07          & 0.712                          & 95.27          & 0.142          & 0.420          &  & 119.09         & 0.288          & 0.551          \\
OTAvatar \cite{ma2023otavatar}          & 30.67                          & 0.654                          & 36.47                          & 0.133                          & 0.226                          &  & 50.37                          & 0.210                          & 0.312                          &                      & 31.23          & 0.695                          & 59.42          & 0.158          & 0.331          &  & 73.30          & 0.304          & 0.492          \\
Real3DPortrait \cite{ye2024real3d}      & \underline{31.62} & 0.712                          & 33.26                          & 0.145                          & 0.163                          &  & 51.36                          & 0.256                          & 0.293                          &                      & \underline{31.54}          & 0.667                          & 79.09          & \underline{0.139}          & \underline{0.325}          &  & 88.91          & \underline{0.262}          & \underline{0.447}          \\
Portrait4D-v2 \cite{deng2024portrait4d} & 30.12                          & \textbf{0.790}                 & 36.57                          & 0.111                          & 0.158                          &  & \underline{42.82} & 0.238                          & \underline{0.248} &                      & 30.29          & \textbf{0.789}                 & \underline{54.95}          & 0.182          & 0.367          &  & \underline{63.97}          & 0.275          & 0.522          \\ \cline{1-20} 
\textbf{Ours}                                            & \textbf{32.40}                 & \underline{0.786} & \textbf{18.95}                 & \textbf{0.102}                 & \textbf{0.129}                 &  & \textbf{27.85}                 & \textbf{0.223}                 & \textbf{0.241}                 &                      & \textbf{31.98} & \underline{0.773} & \textbf{51.35} & \textbf{0.125} & \textbf{0.298} &  & \textbf{57.82} & \textbf{0.210} & \textbf{0.374}
\\ 
\Xhline{5pt}
\end{tabular}%
}
\vspace{-3mm}
\end{table*}
\renewcommand{\arraystretch}{1.0}

\subsection{Symmetric Gaussian Prediction}
\label{sec:appear}
A straightforward approach to generate complete Gaussian parameters would be to concatenate the driven 3D point cloud $[\mathbf{P}_{d}; \mathbf{P}_{d}^{s}]$ with the identity features $\mathbf{F}$ and use a neural network to directly regress the Gaussian parameters. 
However, the symmetric portion of the point cloud represents regions that are not visible in the source image, making it difficult to acquire sufficient information to predict the corresponding Gaussian parameters.
To address this challenge, we propose the Symmetric Gaussian Prediction module, which utilizes a two-stage approach for generating Gaussian parameters.
First, the Gaussian parameters for visible regions are predicted.
Subsequently, the facial symmetry prior is leveraged, along with the previously generated visible Gaussian parameters, to predict those for the symmetric, invisible regions. 
This approach mitigates the information deficiency for invisible regions and enhances the completeness of the Gaussian parameters.
Specifically, we first use the point cloud $\mathbf{P}_d$ and identity encoding $\mathbf{F}$ as inputs to the Gaussian Decoder $D_{gs}$, generating Gaussian parameters for the visible regions, denoted as $\mathcal{G}_d$.
Next, considering the content similarity between the left and right facial regions, we use parameters from the visible regions as a foundation to infer the Gaussian parameters for the invisible regions, which reduces the difficulty of directly predicting parameters for these unseen areas. Specifically we use the Sym-Gaussian Decoder $D_{gs}^{s}$, which takes the generated Gaussian parameters $\mathcal{G}_d$, the identity encoding $\mathbf{F}$ and the symmetric point cloud $\mathbf{P}_d^{s}$ as inputs to predict the symmetric Gaussian parameters $\mathcal{G}_d^{s}$ for the invisible regions.
We concatenate the two sets of Gaussian parameters to obtain the complete Gaussian parameters $\mathcal{G}=[\mathcal{G}_d; \mathcal{G}_d^{s}]$.
Notably, decoders $D_{gs}$ and $D_{gs}^s$ directly use the input dynamic point cloud as the Gaussian position output, while the scaling, rotation, color, and opacity parameters are generated by the network.
Finally, to enhance the detailed rendering capabilities of the Gaussian representation, we apply a parent-child Gaussian densification process \cite{shen2024pixel}, where each Gaussian point in $\mathcal{G}$ generates a child node.
The parent and child nodes together form the dense Gaussian representation $\mathcal{G}_{den}$.
This process can be formulated as follows:
\begin{equation} 
\label{eq:gs1} 
\mathcal{G}_d = D_{gs}(\mathbf{F}, \mathbf{P}_d)
\end{equation}
\vspace{-3mm}
\begin{equation} 
\label{eq:gs2} 
\mathcal{G}_d^{s} = D_{gs}^{s}(\mathcal{G}_d, \mathbf{F}, \mathbf{P}_d^{s})
\end{equation}
\vspace{-2mm}
\begin{equation} 
\label{eq:gs3} 
\mathcal{G}_{den} = Densify([\mathcal{G}_d; \mathcal{G}_d^{s}])
\end{equation}

\subsection{Loss Functions}
\label{sec:loss}
During training, we synthesize both talking face image $\mathbf{I}_{c}^{h}$ and $\mathbf{I}_{tgt}^{h}$ which are rendered before and after densification. Several loss functions are utilized to ensure the similarity between synthesized image and ground truth image $\mathbf{I}_{GT}^{h}$, including L1 loss $\mathcal{L}_1$, SSIM loss $\mathcal{L}_{ssim}$ \cite{wang2004image}, and perceptual loss $\mathcal{L}_{p}$ \cite{johnson2016perceptual}. The overall training objective is as follows: 
\begin{align} \label{eq:loss} 
    \mathcal{L} = \mathcal{L}_1(\mathbf{I}_{c}^{h}, \mathbf{I}_{GT}^{h}) + \mathcal{L}_1(\mathbf{I}_{tgt}^{h}, \mathbf{I}_{GT}^{h}) +
\notag \\
 \lambda_p [\mathcal{L}_{p}(\mathbf{I}_{c}^{h}, \mathbf{I}_{GT}^{h}) + \mathcal{L}_{p}(\mathbf{I}_{tgt}^{h}, \mathbf{I}_{GT}^{h})] +
\notag \\
\lambda_{ssim} \mathcal{L}_{ssim}(\mathbf{I}_{tgt}^{h}, \mathbf{I}_{GT}^{h}) 
\end{align}
% %
 with $\lambda_p=0.01, \lambda_{ssim}=0.2$.

%% file: sec/4_exp.tex
\section{Experiment}
%
% 实验设置

\begin{figure*}[t]
\vspace{-3mm}
\centering
\includegraphics[width=\textwidth]{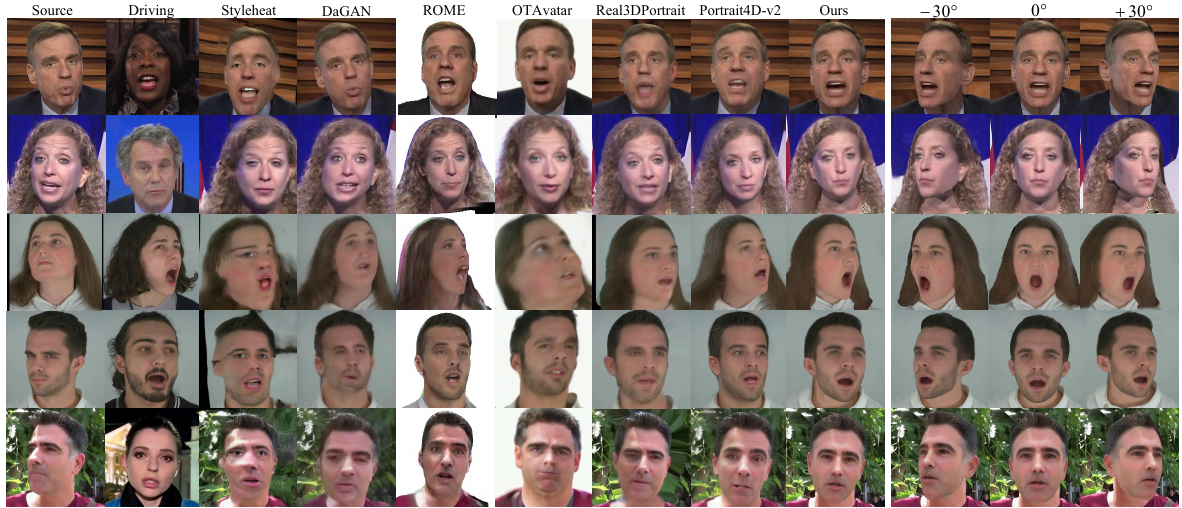}
\vspace{-7mm}
\caption{Qualitative comparisons with previous video-driven methods on the HDTF \cite{zhang2021flow} and NeRSemble-Mono \cite{kirschstein2023NeRSemble} dataset.
%
% The first three rows show the cross-identity driving results on the HDTF dataset, while the last two rows present the results on the NeRSemble dataset. 
The first two rows show the cross-identity driving results on the HDTF dataset, while the third and fourth rows present the results on the NeRSemble-Mono dataset. The last row shows the results of in-the-wild data.
To demonstrate the multi-view consistency of our generated results, the last three columns display the fixed viewpoints at $-30^\circ$, $0^\circ$ and $+30^\circ$.}
\label{fig:video-driven}
\vspace{-4mm}
\end{figure*}

\subsection{Experimental Setups}
\textbf{Datasets.}
% 数据集
We utilized the HDTF \cite{zhang2021flow} and NeRSemble \cite{kirschstein2023NeRSemble} datasets for training.
After removing unstable footage, occlusions, and challenging lighting, approximately 400 video clips remained from HDTF.
Unlike methods \cite{zheng2024headgap,xu20253d} that use NeRSemble for multi-view supervision, we constructed monocular training pairs (source and driving images from the same video), resulting in around 300 video clips, denoted as NeRSemble-Mono.
For preprocessing, we followed AD-NeRF \cite{guo2021ad} to crop video clips to 512×512 and estimate 3DMM pose parameters, then used Deep3DFaceReconstruction \cite{deng2019accurate} to estimate 3DMM expression coefficients for each frame.
For the HDTF and NeRSemble-Mono datasets, we use 80\% for training and the remaining 20\% for testing. To further assess the model's generalization ability, we collected in-the-wild talking videos for additional analysis.

\noindent
\textbf{Evaluation Metrics.}
% 对比指标
We evaluate the methods in video-driven and audio-driven talking head generation, both using a single input image as the identity reference; the former employs a sequence of images, further divided into self-reenactment and cross-reenactment, while the latter relies on audio. 
To assess image quality, we employed metrics including PSNR (Peak Signal-to-Noise Ratio), SSIM (Structural Similarity Index Measure) \cite{wang2004image}, and FID (Frechet Inception Distance) \cite{heusel2017gans}. 
For the video-driven methods, we evaluate expression transition and pose reconstruction using AED (average expression distance)\cite{lin20223d} and APD (average pose distance)\cite{lin20223d}, measured by a 3DMM estimator. 
For the audio-driven methods, we assess lip synchronization with LMD (Landmark Distance)\cite{chen2018lip} and audio-lip synchronization accuracy using LSE-C (Lip Sync Error Confidence)\cite{prajwal2020lip} and LSE-D (Lip Sync Error Distance)\cite{prajwal2020lip}.

\noindent
\textbf{Implementation Details.}
During training, we freeze the Expression Encoder~\cite{zhang2023sadtalker} and monocular depth estimation network \cite{fu2025geowizard}. The model is optimized using the Adam optimizer \cite{kingma2014adam} with a batch size of 1 and a learning rate of $1e^{-4}$. The entire model, excluding the Inpainter, is trained for 1000K steps on a single RTX 4090 GPU, taking approximately 4 days, while the Inpainter is trained separately for 100K steps in under 8 hours. The final model achieves an inference speed exceeding 40 FPS on an RTX 4090.

\begin{figure*}[t]
\vspace{-3mm}
\centering
\includegraphics[width=\textwidth]{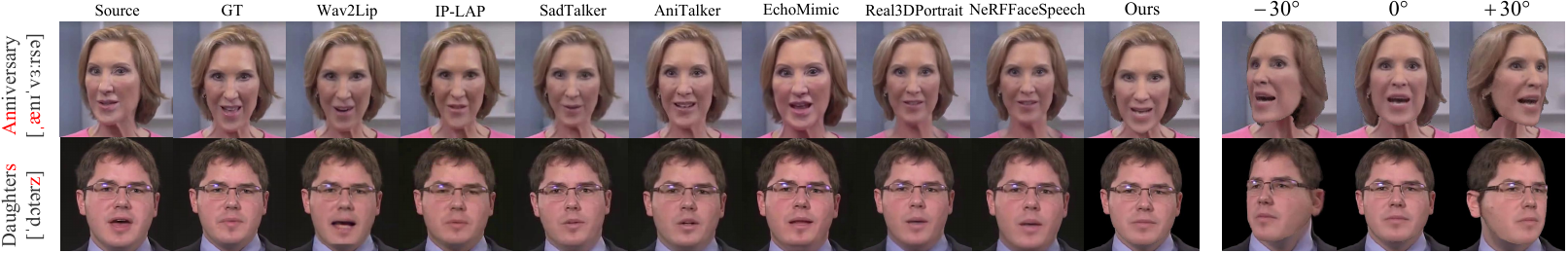}
\vspace{-6mm}
\caption{Qualitative comparisons with previous audio-driven methods on the HDTF \cite{zhang2021flow} dataset.
The last three columns display the fixed viewpoints at $-30^\circ$, $0^\circ$ and $+30^\circ$.}
\label{fig:audio-driven}
\vspace{-3mm}
\end{figure*}

\subsection{Video Driven Talking Head Generation}
\textbf{Quantitative Results.}
We compare our method with state-of-the-art video-driven approaches, including 2D methods like Styleheat \cite{yin2022styleheat} and DaGAN \cite{hong2022depth}, as well as 3D methods like ROME \cite{khakhulin2022realistic}, OTAvatar \cite{ma2023otavatar}, Real3DPortrait \cite{ye2024real3d}, and Portrait4D-v2 \cite{deng2024portrait4d}. Results in Table \ref{tab:video-hdtf} show our method achieves superior PSNR and FID scores in self-reenactment, confirming its effectiveness in high-quality texture generation and identity preservation. Our SSIM score ranks second, slightly behind Portrait4D-v2, highlighting our model's strength in reconstructing facial structure. In cross-reenactment, our approach maintains strong visual quality. Additionally, our method surpasses others on AED and APD metrics in both self- and cross-reenactment, demonstrating precise control of pose and expressions.

\noindent
\textbf{Qualitative Results.}
Figure \ref{fig:video-driven} shows the qualitative results of cross-reenactment between different video-driven methods.
These results show that we are able to effectively recover regions that are invisible in the source image while maintaining strong multi-view consistency. 
On one hand, the explicit decoupling of attributes in 3DGS allows for individual control over geometry, appearance, and pose. On the other hand, our in-depth exploration of facial information enables us to obtain relatively comprehensive Gaussian parameters even from a single reference image.
For example, we observe that our method can generate the ear that is not visible in the source image while maintaining high quality.
Additionally, our method provides better control over facial pose and expression than the competing approaches.
Lastly, we also present the results of our method on in-the-wild data. Despite using less than one-tenth of the training data compared to other methods \cite{ye2024real3d,deng2024portrait4d}, our approach still demonstrates competitive performance.

\renewcommand{\arraystretch}{1.15}
\begin{table}[h]
\fontsize{30}{40}\selectfont
\caption{Quantitative comparisons with the state-of-the-art audio-driven methods on the HDTF dataset \cite{zhang2021flow}.
We use \textbf{bold text} to indicate the best results
and \underline{underline} to denote the second-best results.
The methods \cite{prajwal2020lip,zhong2023identity} highlighted against a gray background, utilize the ground truth upper half of the face as conditional input, focusing on the generation of the lower half of the face.
}
\label{tab:audio driven}
\vspace{-3mm}
\resizebox{\columnwidth}{!}{%
\begin{tabular}{ccccccc}
\Xhline{5pt}
Method                                               & PSNR ↑ & SSIM ↑ & FID ↓ & LMD ↓ & LSE-C ↑ & LSE-D ↓ \\ \hline
Ground Truth                                         & N.A.   & 1.000  & 0.00  & 0.00  & 8.25    & 6.87    \\
\rowcolor{gray!10}  % 设置第二行的背景色为浅灰色
Wav2Lip \cite{prajwal2020lip}       & 34.31  & 0.937  & 18.05 & 3.39  & 8.84    & 6.48    \\
\rowcolor{gray!10}  % 设置第三行的背景色为浅灰色
IP-LAP \cite{zhong2023identity}     & 31.67  & 0.874  & 26.62 & 3.30  & 7.07    & 7.89    \\
SadTalker \cite{zhang2023sadtalker} & 28.60  & 0.567  & 32.77 & 4.07  & \underline{7.63}    & \underline{7.42}   \\
AniTalker \cite{liu2024anitalker}   & 30.42  & 0.620  & 25.74 & 3.87  & 6.12    & 8.23    \\
EchoMimic \cite{chen2024echomimic}  & 29.65  & 0.655  & \underline{22.33} & 3.79  & 6.26    & 8.77    \\
Real3DPortrait \cite{ye2024real3d}                                      & \underline{30.66}  & \underline{0.689} & 39.10 & \textbf{3.61}  & 6.57    & 8.01    \\
NeRFFaceSpeech \cite{kim2024nerffacespeech}                                      & 28.52  & 0.676  & 37.89 & \underline{3.73}  & 6.35    & 8.57    \\ \hline
Ours                                                 & \textbf{30.92}  & \textbf{0.731}  & \textbf{18.73} & 3.75  & \textbf{7.68}    & \textbf{6.91}    \\ \Xhline{5pt}
\end{tabular}%
}
\vspace{-4mm}
\end{table}
\renewcommand{\arraystretch}{1}

\subsection{Audio Driven Talking Head Generation}

\noindent
\textbf{Quantitative Results.}
We compare our method with existing state-of-the-art audio-driven approaches, including 2D image generator-based methods such as Wav2Lip \cite{prajwal2020lip}, IP-LAP \cite{zhong2023identity}, SadTalker \cite{zhang2023sadtalker}, AniTalker \cite{liu2024anitalker}, and EchoMimic \cite{chen2024echomimic}, as well as 3D rendering-based methods like Real3DPortrait \cite{ye2024real3d} and NeRFFaceSpeech \cite{kim2024nerffacespeech}.
Quantitative Evaluation on image quality and audio-lip synchronization is shown in Table \ref{tab:audio driven}, which indicates that our method reaches excellent performance across all image quality assessment metrics compared with leading one-shot audio-driven talking face generation methods.
%
% By employing point cloud densification to address sparsity issues, our method generates videos with a more realistic appearance, accurately replicating the actual scenes. 
%
Regarding lip synchronization, although Wav2Lip performs better on the LSE-C and LSE-D metrics owing to its joint training with SyncNet \cite{chung2017out}, our method remains superior to other fully talking-head generation methods in these metrics. 
Furthermore, our method achieved competitive LMD scores, very close to the second-best method, NeRFFaceSpeech \cite{kim2024nerffacespeech}, indicating its ability to produce convincingly synchronized lip movements with the corresponding phonetic sounds.

\noindent
\textbf{Qualitative Results.}
To qualitatively evaluate different audio-driven methods, we provide sampled images of generated talking face videos in Figure \ref{fig:audio-driven}. 
Specifically, the source image and the ground truth (GT) are shown in the first and second columns of the figure, followed by synthesized images from different methods. 
As observed, beyond the superior visual quality and identity preservation, our method achieves mouth shapes and poses closer to the GT, demonstrating enhanced pose control and lip synchronization capabilities.
% %
%
% %

% %
%

%
%
%
%
%
\renewcommand{\arraystretch}{1.0}
\renewcommand{\arraystretch}{1.2}
\begin{table}[]
\caption{Ablation results on the HDTF dataset in self-reenactment setting.
The first three rows show the ablation results of the Depth-Aware Symmetric Geometry Reconstruction (DSGR) module, while the fourth to sixth rows present the ablation results of the Symmetric Gaussian Prediction (SGP) module.}
\label{tab:ab-study}
\vspace{-2mm}
\fontsize{30}{40}\selectfont
\resizebox{\columnwidth}{!}{%
\begin{tabular}{clccccc}
\Xhline{5pt}
Module & Method                                   & PSNR ↑         & SSIM ↑         & FID ↓          & AED↓           & APD↓           \\ \hline
\multirow{3}{*}{DSGR} & w/o Geo. Refine.                      & 29.98          & 0.705          & 24.82          & 0.157          & 0.232          \\
                                     & w/o Sym.                            & 31.24          & 0.760          & 20.23          & 0.112          & 0.147          \\
                                     & w/o Gauss. Filter                & 29.56          & 0.724          & 27.42          & 0.148          & 0.180          \\ \hline
\multirow{4}{*}{SGP}                  & w/o Pointclouds Input & 30.78          & 0.776          & 22.05          & 0.124          & 0.134          \\
                                     & w/o Sym. Gauss. Dec.                       & 32.12          & 0.761          & 19.74          & 0.116          & 0.151          \\
                                     & w/o $\mathcal{G}_d$ in Sym. Gauss. Dec.    & 31.68          & 0.764          & 19.36          & 0.109          & 0.142          \\
                                     & w/o $Densify$                            & 30.32          & 0.778          & 21.58          & 0.138          & 0.133          \\ \hline
                     & Ours                                     & \textbf{32.40} & \textbf{0.786} & \textbf{18.95} & \textbf{0.102} & \textbf{0.129}
                     \\
                     \Xhline{5pt}
\end{tabular}%
}\vspace{-3mm}
\end{table}
\renewcommand{\arraystretch}{1.0}
\begin{figure}[h]
\centering
\includegraphics[width=0.48\textwidth]{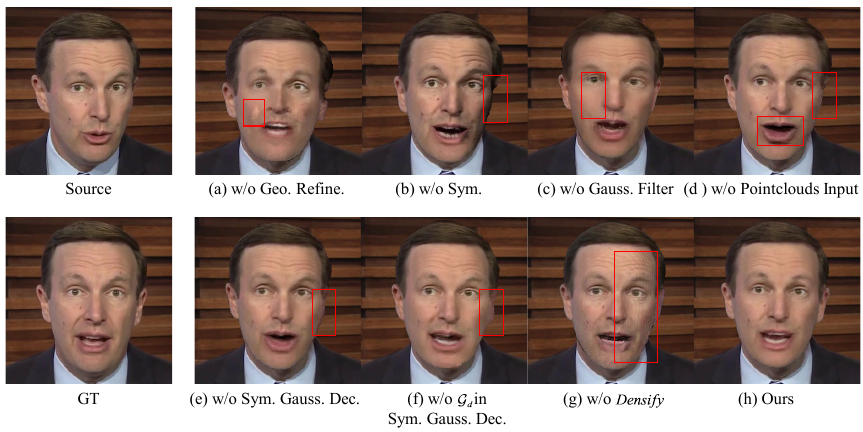}
\vspace{-6.5mm}
\caption{Visualization of ablation study results. 
The first image shows the source image and the ground truth. 
Figures (a) to (h) correspond to rows one through eight in the Table \ref{tab:ab-study}.
The differences between images can be observed in greater detail by zooming in to 250\%.}
\label{fig:ab}
\vspace{-3mm}
\end{figure}
\subsection{Ablation Study}
\label{sec:ab-study}
As discussed above, the use of symmetry is crucial for obtaining complete geometry and Gaussian parameters. 
Therefore, in the following section, we focus on the role of facial symmetry priors in the Depth-Aware Symmetric Geometry Reconstruction (DSGR) and Symmetric Gaussian Prediction (SGP) modules, as well as ablate specific design choices within each module to further understand their impact on the overall performance.
The quantitative results for the self-reenactment video-driven scenario are presented in Table \ref{tab:ab-study}, with the corresponding visual results shown in Figure \ref{fig:ab}.

\subsubsection{Depth-Aware Symmetric Geometry Reconstruction}

\noindent
\textbf{Effect of Symmetric Operation}. To measure the importance of the symmetric operation, we removed it from the DSGR module (w/o Sym.). 
Since monocular depth estimation cannot provide depth information for the facial regions not visible in the source image, using only the point cloud obtained from depth estimation to represent the facial geometry results in incomplete modeling of the invisible regions. 
As shown in Figure \ref{fig:ab} (b), the model fails to reconstruct the left ear that is occluded in the source image.
To address this issue, we utilized symmetry to mirror the point cloud and fill in the missing regions.
However, directly mirroring the point cloud can introduce significant overlap between the original and mirrored point clouds.
Without filtering these overlapping regions (w/o Gauss. Filter), the two point clouds interfere with each other, leading to a loss of details in the generated images, as shown in Figure \ref{fig:ab} (c). 
Table \ref{tab:ab-study} also shows that without filtering to alleviate point cloud overlap, all metrics degrade significantly.
\begin{figure}[t]
\hspace{-2mm}
\includegraphics[width=0.5\textwidth]{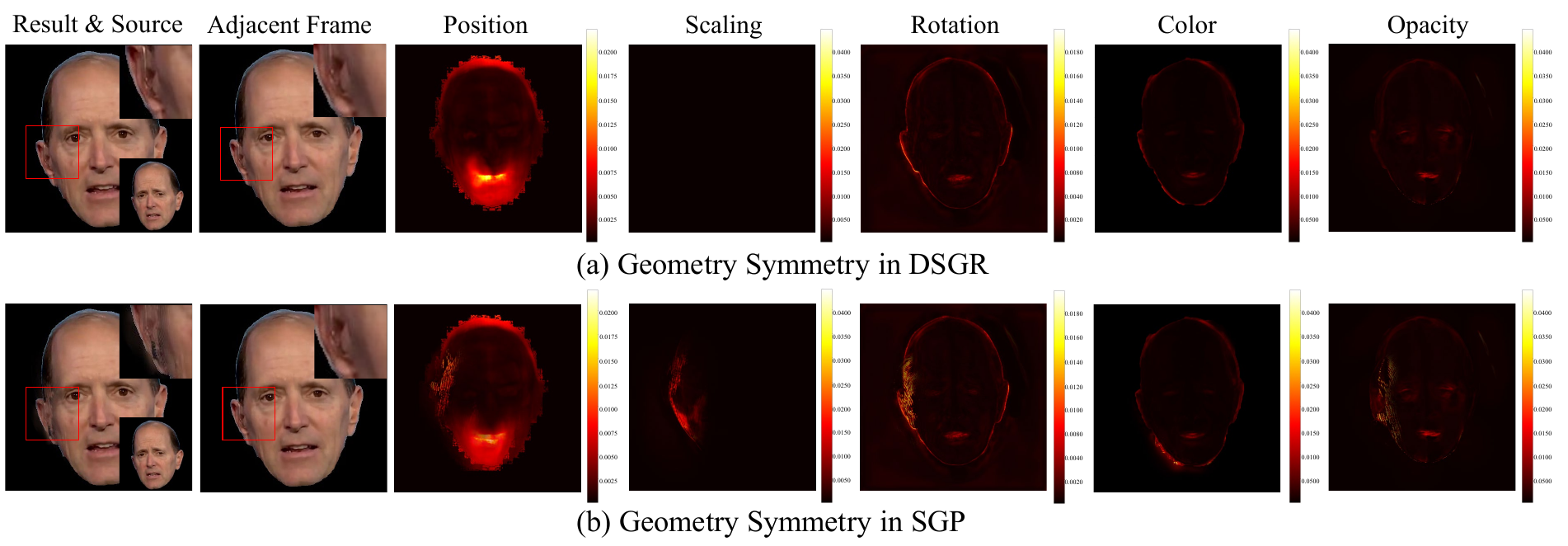}
\vspace{-6mm}
\caption{
Impact of placing geometric symmetry in the DSGR or SGP module. The visualization shows two consecutive frames along with the variance of Gaussian parameters across all frames in the generated video. Differences between images can be examined in greater detail by zooming in to 350\%.
}
\label{fig:consis}
\vspace{-2mm}
\end{figure}
\noindent
\textbf{Effect of Refinement Network}.
When this refinement network is removed (w/o Geo. Refine.), errors in the initial depth estimation propagate, affecting the prediction of all Gaussian parameters and resulting in degraded pose and appearance in the generated images, as shown in Figure \ref{fig:ab} (a).
\subsubsection{Symmetric Gaussian Prediction}
\noindent
\textbf{Effect of Two-stage Generation Network for Symmetric Prior}.
We first replace the geometric information input in the SGP module with expression features directly (w/o Pointclouds Input).
Without geometric awareness for each pixel, the model fails to reconstruct the invisible regions, and the visual quality in high-frequency motion areas, such as the mouth, also deteriorates (Figure \ref{fig:ab} (d)).
We further investigate the generation strategy for symmetric Gaussian parameters. 
As shown in Figure \ref{fig:ab} (e)(f), when removing the Sym-Gaussian Decoder and using a single network to generate all Gaussian parameters simultaneously (w/o Sym. Gauss. Dec.), or using two separate networks to predict each part of the Gaussian parameters without providing the visible region's Gaussian parameters as input to the Sym-Gaussian Decoder (w/o $\mathcal{G}_d$ in Sym. Gauss. Dec.), they fail to generate well in the occluded regions of the source image. 
Due to the limited input information, directly learning parameters for occluded regions poses significant challenges.
Using parameters from visible regions as guidance leverages their correlation, reducing the difficulty of prediction.

\noindent
\textbf{Effect of Densify Strategy}. Removing the densification operation reduces generation quality, as demonstrated in Table 3 for quantitative results and visualized in Figure \ref{fig:ab} (g), where the sparsity of points results in noticeable Moiré-like patterns in facial regions.

\renewcommand{\arraystretch}{1.3}
\begin{table}
   \caption{User study. We use \textbf{bold text} to indicate the best results
and \underline{underline} to denote the second-best results. 
}
  \label{tab:user study}
  \vspace{-2mm}
\fontsize{30}{40}\selectfont
\resizebox{\columnwidth}{!}{
  \begin{tabular}{ccccccccc}
  \Xhline{5pt}
    \addlinespace[1pt]   
    \multirow{2}{*}{Method} & \multicolumn{4}{c}{Video-driven} & \multicolumn{4}{c}{Audio-driven} \\
    \cmidrule(lr){2-5} \cmidrule(lr){6-9}
    &Ours &Portrait4D-v2 &DaGAN &Others &Ours  &EchoMimic &Wav2Lip &Others  \\ 
    \hline
    ID. Pres. & \textbf{45.0\%} & \underline{35.0\%} & 5.0\% &15.0\% &\textbf{37.5\%} &\underline{25.0}\% &17.5\% &20.0\% \\
    Vis. Quality &\textbf{42.5\%} &\underline{32.5\%} &5.0\% &20.0\% &\textbf{40.0\%} &{12.5\%} &\underline{25.0\%} &22.5\% \\
    Lip Sync. &\textbf{32.5\%} &\underline{27.5\%} &12.5\% &27.5\% &\underline{35.0\%} &12.5\% &\textbf{40.0\%} &12.5\% \\
    \Xhline{5pt}
  \end{tabular}}
\vspace{-7mm}
\end{table}
\renewcommand{\arraystretch}{1}

\subsubsection{Impact of Geometry Symmetry Placement}
\label{sec:consis}
%
% %
%
This section explores the impact of applying geometric symmetry in the DSGR versus the SGP module.
When symmetry is applied in the SGP module, the driven point cloud is influenced by driving expressions, leading to inconsistency in the mirrored geometry.
This inconsistency causes significant variance in Gaussian parameters across frames, resulting in unstable visual output (Figure \ref{fig:consis} (b)).
To avoid this, we apply symmetry within the DSGR module to construct a consistent facial geometry before deformation.
This approach ensures stable parameter prediction across frames and achieves a consistent facial appearance in the generated video (Figure \ref{fig:consis} (a)).

\subsection{User Study}
We conducted a user study with 40 participants to evaluate the quality of talking head animations generated by different methods, using identical audio or video inputs for 30 clips. 
Participants were asked to evaluate three aspects: Identity Preservation, Visual Quality (intelligibility and naturalness), and Lip Synchronization.
The questionnaire format is similar to KMTalk~\cite{xu2024kmtalk}.
As summarized in Table \ref{tab:user study}, our method achieves the best results in both identity preservation and visual quality.
For lip synchronization, our method ranks highest for video-driven inputs and second only to Wav2Lip \cite{prajwal2020lip} for audio-driven inputs.

\subsection{Robustness of Asymmetric Heads}
\begin{figure}[h]
\vspace{-2mm}
\setlength{\abovecaptionskip}{0.cm}
    \centering
    \includegraphics[width=1\linewidth]{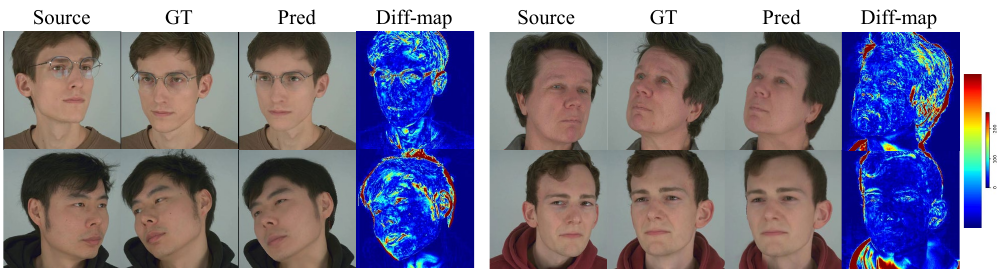}
    \caption{
Qualitative results of asymmetric hairstyles.
The Diff-map shows the squared difference between the GT and Pred.}
    \label{fig:sym_ab}
\end{figure}

The SGP module generates Gaussian parameters for the invisible region based on the visible area guidance and symmetric point cloud, rather than simply copying parameters. Thus it can handle some head asymmetries (low reconstruction errors of asymmetric hairs in Fig.~\ref{fig:sym_ab}).

\vspace{-5mm}

%% file: sec/5_conclusion.tex
\section{Conclusion}
\label{sec:con}
We introduce MGGTalk, a novel framework that utilizes generalizable 3D Gaussian Splatting (3DGS) for one-shot talking head generation, trained exclusively on monocular data. 
The framework integrates facial depth geometry and symmetry priors across two core modules, enabling efficient prediction of 3DGS parameters and facilitating high-quality, multi-view consistent rendering.
Evaluated on both audio-driven and video-driven tasks, MGGTalk demonstrates significant improvements in both quantitative and qualitative metrics compared to baseline methods. 

%% file: sec/X_suppl.tex
\clearpage
\setcounter{page}{1}
\maketitlesupplementary

\section{Limitations and Future Work}
\label{sec:limitations}
Despite the progress achieved in our talking head animation approach MGGTalk, several limitations warrant further investigation. We identify four primary areas for improvement: (1) Unnatural connection among the head, neck, and torso. Future work could employ a unified model of head, neck, and torso to enhance realism in their transitions and overall alignment. (2) Insufficient utilization of video information to improve context consistency. Incorporating multi-frame constraints during training could better estimate identity-specific shapes and maintain temporal coherence, thereby strengthening the naturalness of generated outputs~\cite{tewari2021learning}. (3) Potential inaccuracies in single-view depth estimation. Errors in depth estimates can compromise 3D modeling accuracy; adopting more robust approaches—such as DPT~\cite{ranftl2021vision} or Sapiens~\cite{khirodkar2024sapiens}—may substantially improve reconstruction fidelity. (4) Unnatural results under severe asymmetry or challenging illumination. Complex lighting conditions can lead to unrealistic renderings, suggesting the need for illumination-aware control that adapts generation to diverse lighting environments. Addressing these limitations will further refine the quality, realism, and robustness of Talking Head Animation methods.

\section{Ethics Consideration}
\label{sec:ethics}
The proposed talking head animation method is primarily intended for applications in virtual communication and entertainment. Nonetheless, it may raise ethical and legal concerns if exploited for deceptive or harmful purposes by malicious actors. To mitigate these risks, it is essential to establish clear ethical guidelines and responsible usage practices that explicitly prohibit misuse. By doing so, we can help ensure that this technology is employed in a manner that promotes beneficial applications while minimizing potential harm.

\section{Preliminary of 3DGS}
\label{sec:pre}
3D Gaussian Splatting (3DGS) \cite{kerbl20233d} utilizes anisotropic 3D Gaussians as geometric primitives to learn an explicit 3D representation. The geometry of each 3D Gaussian is defined as follows:
\begin{equation} \label{eq:SSG} 
g(x) = e^{( -\frac{1}{2}(x - \mu)^{T} \Sigma^{-1} (x - \mu) )}
\end{equation}
where $\mu \in \mathbb{R}^{3}$ is the center of the Gaussian and $\Sigma \in \mathbb{R}^{3 \times 3}$ is the covariance matrix that defines its shape and size. 
The covariance matrix $\Sigma$ can be further decomposed into $\Sigma = RSS^{T}R^{T}$, where S denotes a scaling matrix determined by a scaling vector $s \in \mathbb{R}^{3}$, while R indicates a rotation matrix defined by a quaternion $r \in \mathbb{R}^{4}$. Additionally, each Gaussian has an opacity value $o \in \mathbb{R}$ which determines its visibility, and a color feature defined by $c \in \mathbb{R}^{12}$. 
%SH
%
Collectively, these parameters define each Gaussian as $\mathcal{G} = \{\mu, r, s, o, c\}$. 
Specifically, $\mu$ represents the position parameter of the Gaussian, which will be equivalently referred to as the three-dimensional coordinates of points in the Gaussian point cloud $\mathbf{P}$ in the subsequent discussion.
\begin{figure}[h]
\centering
\includegraphics[width=0.45\textwidth]{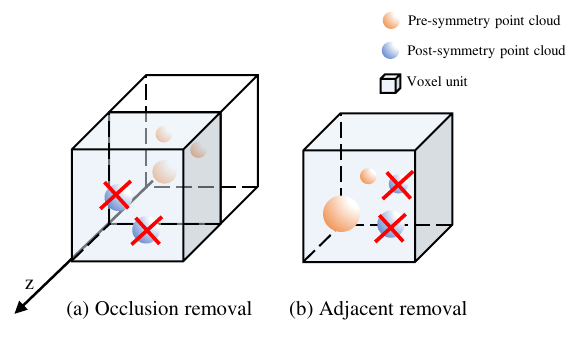}
\caption{Visualization of voxel filter rules. (a) illustrates occlusion removal along the z-axis, while (b) shows adjacent removal.}
\label{fig:voxel}
\end{figure}
\section{Implementation Details}
\subsection{Voxel-Based Filter}
\label{sec:voxel}
We propose a voxel-based filter $\mathcal{F}_{voxel}$  to effectively remove occluded or closely overlapping mirrored points from point clouds.
The method voxelizes both the original and mirrored point clouds, then performs two key operations: $z$-axis occlusion detection and neighborhood occupancy checking.
For $z$-axis occlusion, we compute the maximum $z$-values at each $(x,y)$ voxel index in the original point cloud and compare them with mirrored points to discard occluded regions, as shown in Figure \ref{fig:voxel} (a).
In the neighborhood check, we determine if mirrored points fall within the same voxel as the original points, considering them as neighboring points, and thus discarding them, as shown in Figure \ref{fig:voxel} (b). 
This combined approach efficiently retains essential mirrored points while removing those that are occluded or redundant.
\subsection{Motion Deformation}
\label{sec:Deform}
The Motion Deformation module is designed to deform the 3D point cloud $[\mathbf{P}_{f};\mathbf{P}_f^s]$ to synchronize it with the driving audio or driving image.
Unlike directly editing the 3D point cloud using only the driving source, we also incorporate expression information from the source image to reduce the complexity of mapping arbitrary source expressions to arbitrary target expressions.
We use 3DMM reconstruction \cite{deng2019accurate} to extract the source expression basis $\beta_{s}$ from the source image. For the driving input, the driving expression basis $\beta_{d}$ is obtained via 3DMM reconstruction for driving images or an audio-to-expression method \cite{zhang2023sadtalker} for driving audio. The $\mathbf{MLP}$ then drives the source point cloud $[\mathbf{P}_{f};\mathbf{P}_{f}^s]$ to generate the driven point cloud $[\mathbf{P}_{d};\mathbf{P}_{d}^s]$ (Equation \ref{eq:deform2}).
\begin{equation} \label{eq:deform2} 
[\mathbf{P}_d; \mathbf{P}_d^s] = \mathbf{MLP}([\mathbf{P}_f; \mathbf{P}_f^s], \beta_d, \beta_s)
\end{equation}
%
%
% %

% %
%
%
\subsection{Gaussian Decoder}
\label{sec:gs-dec}
The Gaussian Decoder $D_{gs}$ is designed to predict the remaining four Gaussian parameters—scaling $\mathbf{s}$, rotation $\mathbf{r}$, color $\mathbf{c}$, and opacity $\mathbf{o}$—for the visible deformed region point cloud $\mathbf{P}_d$. 
First, the input point cloud is reshaped into the form of a position map with dimensions (3, H, W). 
This map is then concatenated with the identity feature $\mathbf{F}$ extracted from the source image and fed into a UNet-based network to generate the $\mathbf{s}$, $\mathbf{r}$, $\mathbf{c}$, and $\mathbf{o}$. 
Finally, these maps are reshaped back into point cloud format and concatenated to form the complete set of Gaussian parameters $\mathcal{G}_d=\{ \mathbf{P}_d\in \mathbb{R}^{H\cdot W \times 3}, \mathbf{s}\in \mathbb{R}^{H\cdot W \times 3}, \mathbf{r}\in \mathbb{R}^{H\cdot W \times 4},\mathbf{c}\in \mathbb{R}^{H\cdot W \times 12}, \mathbf{o}\in \mathbb{R}^{H\cdot W \times 1} \}$.
\begin{figure*}[t]
\centering
\includegraphics[width=0.85\textwidth]{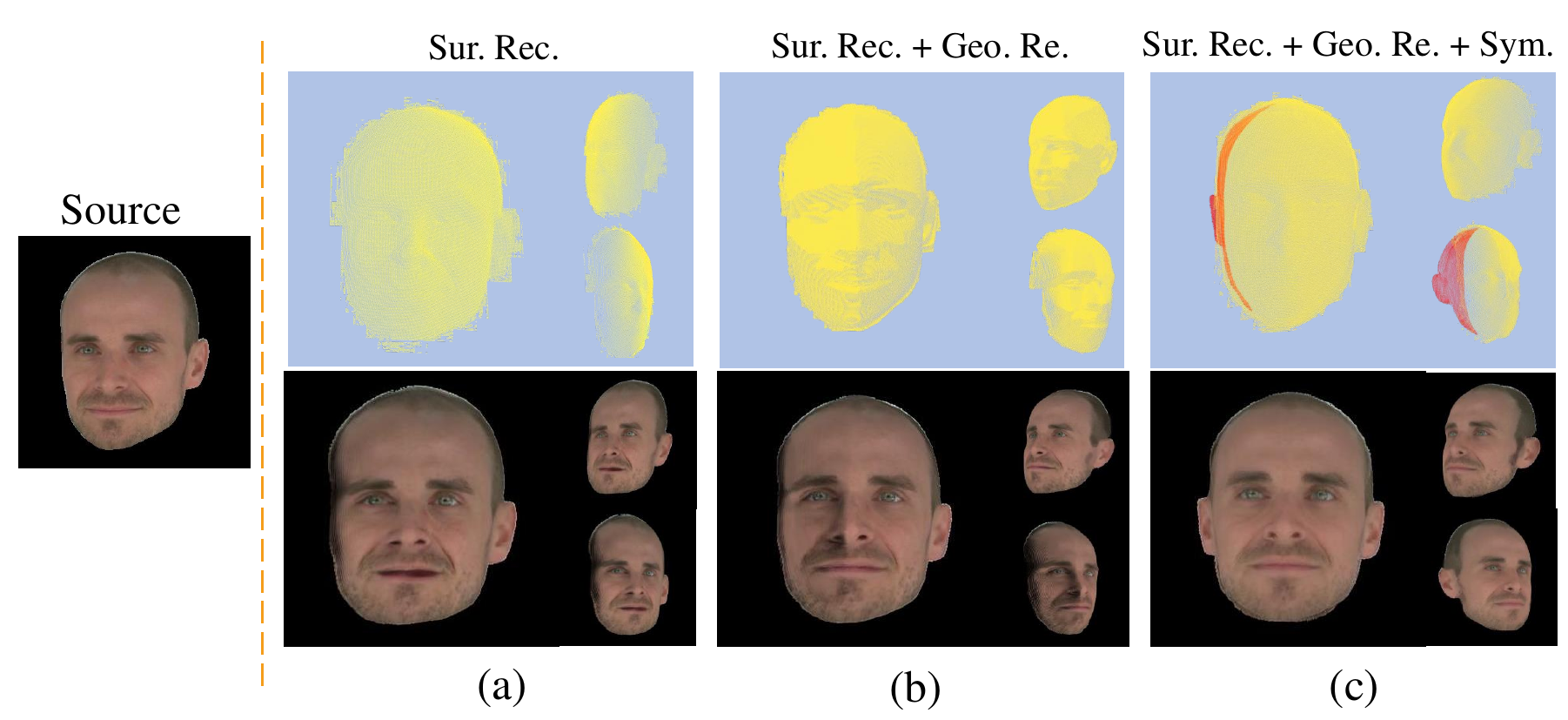}
\caption{Visualization of the Gaussian point cloud construction in DSGR. 
The first row shows Gaussian point clouds obtained by ablating different construction modules, while the second row presents the corresponding rendered images from the SGP module.
The yellow point cloud represents the geometry obtained from depth information, while the red point cloud indicates the symmetric augmentation of the geometry.
Front, left, and right viewpoints are displayed.}
\label{fig:geometry}
\end{figure*}
\subsection{Sym-Gaussian Decoder}
\label{sec:sym-gs-dec}
The Sym-Gaussian Decoder $D_{gs}^s$ is designed to generate Gaussian parameters for the non-visible regions of the point cloud $\mathbf{P}_d^s$. 
Given the challenge of obtaining sufficient information for these regions from the source image alone, facial symmetry priors are introduced as additional guidance. 
Specifically, the previously generated Gaussian parameters for the visible regions $\mathcal{G}_d$, identity features $\mathbf{F}$, and the symmetric point cloud $\mathbf{P}_d^s$ are concatenated and fed as input to a convolutional network to predict the offset relative to the already generated parameters. 
The networks for generating the biases of each Gaussian parameter are denoted as $D_s^s$, $D_r^s$, $D_c^s$, and $D_o^s$, respectively, and the generation process can be expressed as follows:
\begin{equation}
\label{eq:sym-gs} 
\left\{ \begin{array}{l}
	\mathbf{s}^s=\mathbf{s}+D_{s}^{s}\left( \mathbf{F,P}_{d}^{s},\mathbf{s} \right)\\
	\mathbf{r}^s=\mathbf{r}+D_{r}^{s}\left( \mathbf{F,P}_{d}^{s},\mathbf{r} \right)\\
	\mathbf{c}^s=\mathbf{c}+D_{c}^{s}\left( \mathbf{F,P}_{d}^{s},\mathbf{c} \right)\\
	\mathbf{o}^s=\mathbf{o}+D_{o}^{s}\left( \mathbf{F,P}_{d}^{s},\mathbf{o} \right)\\
\end{array} \right. 
\end{equation}
Finally, we obtain the Gaussian parameters $\mathcal{G}_d^s=\{ \mathbf{P}_d^s, \mathbf{s}^s, \mathbf{r}^s,\mathbf{c}^s,\mathbf{o}^s \}$ representing the non-visible facial regions of the source image.
\subsection{Rendering and Inpainting}
\label{sec:render}
We use differentiable rasterization to render the Gaussian parameters $\mathcal{G}_{den}$ from the target viewpoint, resulting in an RGB image $\mathbf{I}^{h}_{tgt}$. 
To stabilize the training process, we additionally render the Gaussian parameters $\mathcal{G}$ before densification into a coarse image $\mathbf{I}^{h}_{c}$.
Subsequently, we inpaint the torso and background regions of $\mathbf{I}^{h}_{tgt}$ using $\mathbf{I}_s^{bg}$, producing the final predicted image $\mathbf{I}_{tgt}$.
Inspired by $\mathrm{S}^3$D-NeRF \cite{li2024s}, we employ a GAN-based network that takes the head image and the torso-background image as inputs to generate a 512 $\times$ 512 composite image.

\section{Additional Results}
To demonstrate the effectiveness of our approach, we provide additional visualizations and experimental results within the context of the video-driven talking head generation task.

\begin{figure}[t]
\centering
\includegraphics[width=0.45\textwidth]{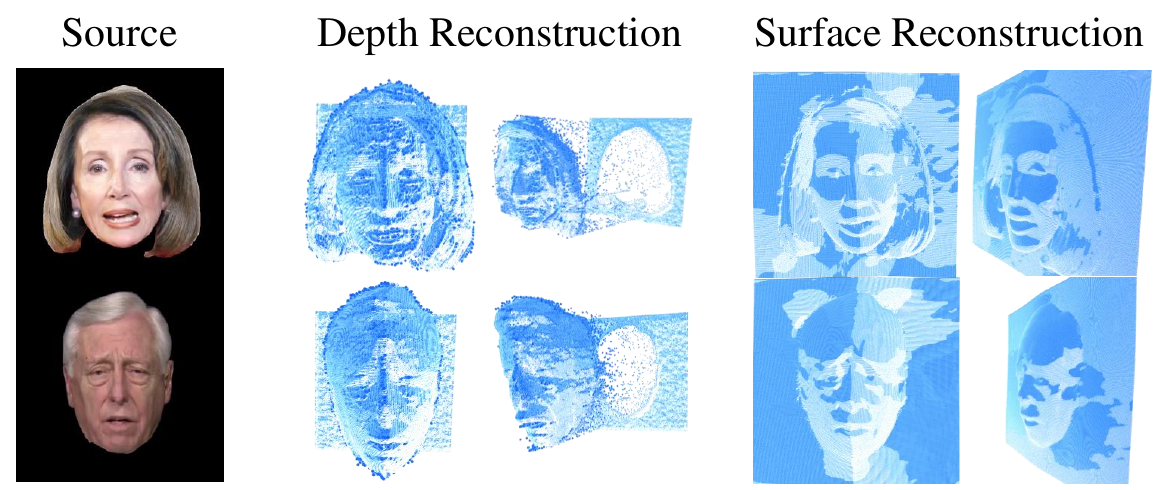}
\caption{Visualization of geometric structures obtained from depth map projection versus surface reconstruction, including both frontal and side views of the 3D point cloud.}
\label{fig:depth}
\end{figure}

\subsection{Visualization of Gaussian Point Cloud Construction in DSGR}
We utilize visualizations in Figure \ref{fig:geometry} to observe the three stages of point cloud construction in DSGR moudle. Initially, we use depth maps combined with normal maps as input for the Surface Reconstruction module, forming the initial facial geometry. Subsequently, a refinement network adjusts the initial construction, and facial symmetry is introduced to supplement the missing geometric structure in the occluded areas of the face. 

For the first stage, the combination of depth and normal maps is critical. As illustrated in Figure \ref{fig:depth}, although the geometry derived directly from the depth map exhibits a stronger sense of three-dimensionality, it is often inaccurate due to monocular depth estimation limitations. For example, the first row shows an exaggerated nose, and the second row an overly sharp chin. Additionally, the geometric continuity of point clouds obtained through depth map back-projection is often inadequate, which hinders network training convergence. To address these inaccuracies, we incorporate normal maps to enhance geometric details. Both depth and normal maps are then used as inputs to the BINI algorithm \cite{cao2022bilateral} for surface reconstruction, producing more continuous and accurate 3D facial point clouds, as shown in Figure \ref{fig:depth} and \ref{fig:geometry} (a), where surface reconstruction achieves smoother geometric continuity without the aforementioned structural inaccuracies.

\renewcommand{\arraystretch}{1.5} 
\begin{table}[t]
\fontsize{30}{40}\selectfont
\caption{Quantitative results of video-driven methods on the CelebV-HQ dataset \cite{zhu2022celebv}.
We use \textbf{bold text} to indicate the best results and \underline{underline} to denote the second-best results.}
\label{tab:celeb}
\resizebox{\columnwidth}{!}{%
\begin{tabular}{cccccccccc}
\Xhline{5pt}   
% \cline{2-10}
\multirow{2}{*}{Methods}                                 & \multicolumn{5}{c}{Self-Reenactment}                                                                                                                               &  & \multicolumn{3}{c}{Cross-Reenactment}                                                            \\ \cline{2-6} \cline{8-10} 
                                                         & PSNR↑                          & SSIM↑                          & FID↓                           & AED↓                           & APD↓                           &  & FID↓                           & AED↓                           & APD↓                           \\ \hline
Styleheat \cite{yin2022styleheat}       & 30.36                          & 0.634                          & 71.57                          & 0.157                          & 0.383                          &  & 83.12                          & 0.224                          & 0.405                          \\
DaGAN \cite{hong2022depth}              & \underline{30.81}                          & 0.626                          & 57.72 & 0.113 & 0.196                          &  & 60.45                          & 0.244 & 0.308                        \\
ROME \cite{khakhulin2022realistic}      & 30.74                          & 0.657                          & 62.66                          & 0.140                          & \underline{0.179} &  & 78.02                          & 0.257                          & 0.283                          \\
OTAvatar \cite{ma2023otavatar}          & 30.37                          & 0.681                          & 50.03                          & 0.136                          & 0.352                          &  & 64.21                          & 0.205                          & 0.371                         \\
Real3DPortrait \cite{ye2024real3d}      & 30.67 & \textbf{0.696}                          & 73.17                         & \underline{0.109}                          & 0.231                          &  & 75.16                          & \textbf{0.191}                          & \textbf{0.254}                          \\
Portrait4D-v2 \cite{deng2024portrait4d} & 29.96                          & 0.613                 & \underline{46.19}                          & 0.112                          & 0.216                          &  & \underline{57.13} & 0.208                          & 0.262 \\ \hline
\textbf{Ours}                                            & \textbf{30.84}                 & \underline{0.683} & \textbf{42.23}                 & \textbf{0.104}                 & \textbf{0.173}                 &  & \textbf{56.43}                 & \underline{0.195}                & \underline{0.256}                 \\
\Xhline{5pt}                                &                                &                                &                                &                                &                                &  &                                &                                &                               
\end{tabular}%
}
\end{table}
\renewcommand{\arraystretch}{1.0} 

The refinement network, detailed in the next phase, further adjusts the geometry to correct any residual inaccuracies, as illustrated in the Figure \ref{fig:geometry} (b). Although the initial point cloud constructed from depth and normal maps provides a good foundation, it may appear flat and fail to accurately represent the 3D facial structure. The refinement module effectively addresses these issues.

Finally, the application of symmetry plays a crucial role in reconstructing occluded regions of the face, which are often left incomplete in the initial stages. As shown in Figure \ref{fig:geometry} (c), The symmetry approach fills these gaps, ensuring a more comprehensive and accurate representation of the facial geometry across the entire point cloud.

\begin{figure*}[t]
\centering
% \vspace{-50mm}
\includegraphics[width=\textwidth]{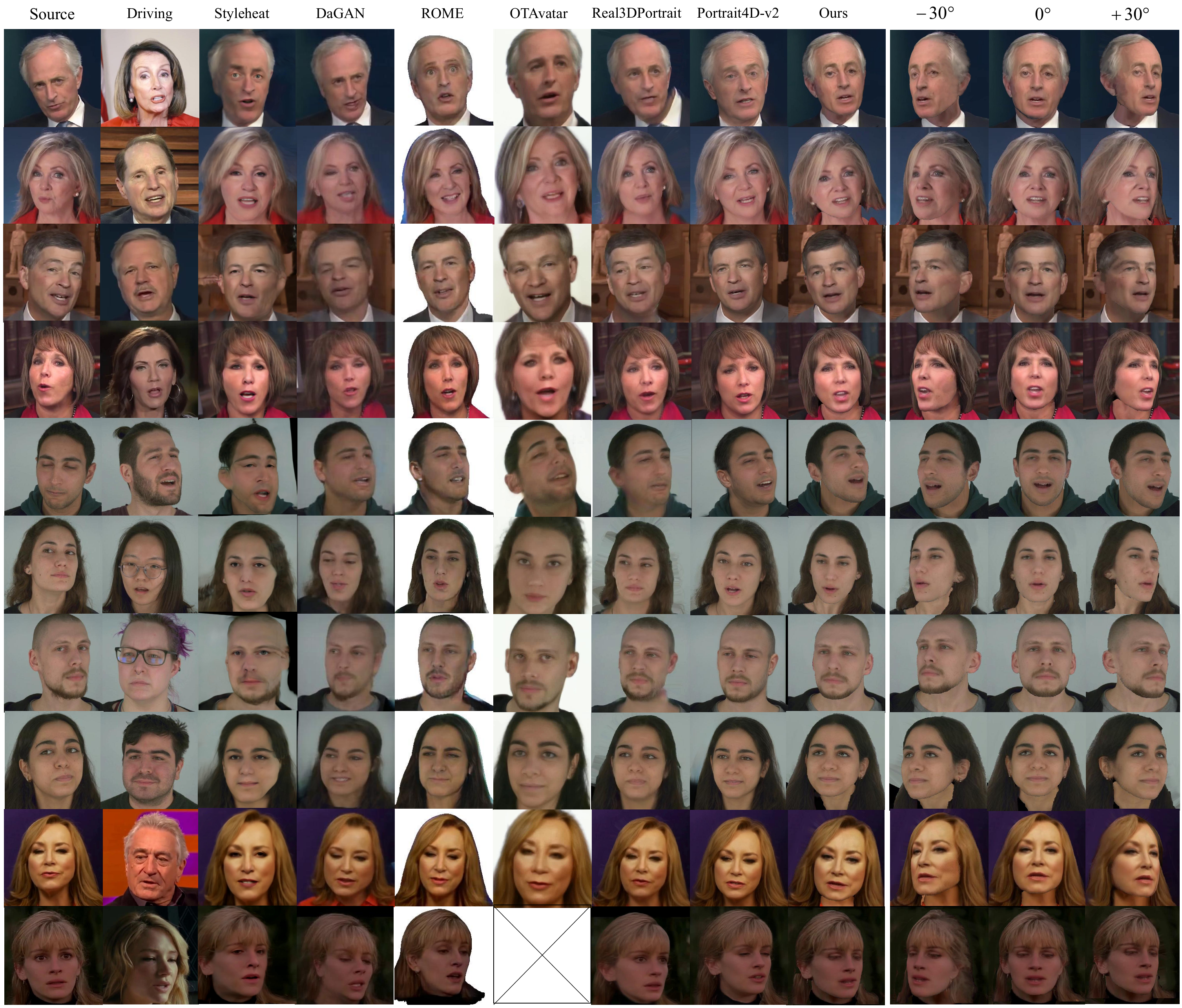}
\vspace{-7mm}
\caption{Qualitative comparisons with previous video-driven methods on the HDTF \cite{zhang2021flow}, NeRSemble-Mono \cite{kirschstein2023NeRSemble} and CelebV-HQ \cite{zhu2022celebv} dataset.
The first four rows show cross-identity driving results on the HDTF dataset, rows five to eight present results on the NeRSemble-Mono dataset, and the final two rows display results from the CelebV-HQ dataset.
To demonstrate the multi-view consistency of our generated results, the last three columns display the fixed viewpoints at $-30^\circ$, $0^\circ$ and $+30^\circ$.}
\label{fig:video-driven-more}
\end{figure*}

\subsection{Additional Results on HDTF and NeRSemble-Mono} 
In Figure \ref{fig:video-driven-more}, we present additional cross-identity reenactment results on the HDTF dataset (first four rows) and the NeRSemble-Mono dataset (rows five to eight). 
The results demonstrate that our method achieves strong identity consistency and 3D coherence, while effectively synchronizing facial expressions and poses with the driving source.
\subsection{Additional Experiments on CelebV-HQ}
%\jiapeng{on which task? audio-driven or video driven}
%
\noindent
\textbf{Experimental Setups.}
To further evaluate the model's performance, we employed an additional dataset, CelebV-HQ \cite{zhu2022celebv}, for quantitative and qualitative experiments on video-driven methods. 
Specifically, no training was conducted on this dataset; instead, 40 video clips were selected for inference. 
Data preprocessing and evaluation metrics were consistent with those used in the main text.
\noindent
\textbf{Quantitative Results.}
Experimental results on the CelebV-HQ dataset are presented in Table \ref{tab:celeb}.
For Self-Reenactment, our method outperforms others in appearance quality metrics (PSNR and FID) and is comparable to Real3DPortrait \cite{ye2024real3d} in SSIM, indicating structural similarity.
For Cross-Reenactment, our approach maintains a lead in FID, demonstrating superior identity preservation. 
Additionally, our AED and APD scores are close to Real3DPortrait \cite{ye2024real3d}, indicating effective control of facial expressions and poses.
\noindent
\textbf{Qualitative Results.}
The visual results on the CelebV-HQ dataset are shown in the last two rows of Figure \ref{fig:video-driven-more}. 
Despite using significantly less training data compared to some methods \cite{ye2024real3d, deng2024portrait4d}, our approach demonstrates competitive performance on unseen, in-the-wild dataset \cite{zhu2022celebv}, maintaining strong 3D consistency as well as effective synchronization of expressions and poses.
\subsection{Further Enhancement of Lip Synchronization}
Our MGGTalk framework already achieves the second-best performance in terms of lip synchronization (LSE-C, LSE-D). As shown in Table \ref{tab:syncnet}, introducing SyncNet~\cite{chung2017out} provides additional performance improvements, suggesting that adopting an audio-based synchronization module can further refine the lip-sync accuracy of our method.

\begin{table}[h]
\fontsize{10}{15}\selectfont
\centering
\caption{
Audio-riven results on HDTF~\cite{zhang2021flow} with the SyncNet~\cite{chung2017out} supervision.
}
\label{tab:syncnet}
\resizebox{0.7\columnwidth}{!}{%
\begin{tabular}{ccc}
\Xhline{1pt}
Method                  & LSE-C↑ & LSE-D↓ \\ \hline
Wav2Lip~\cite{prajwal2020lip}         & 8.84   & 6.48   \\
MGGTalk                 & 7.68   & 6.91   \\
MGGTalk+SyncNet~\cite{chung2017out} & \textbf{8.87}   & \textbf{6.35}   \\ \Xhline{1pt}
\end{tabular}
}
\end{table}

\subsection{A Fairer Comparison with Lip-Sync Methods}
In Table \ref{tab:audio driven} of the main paper, we note that both Wav2Lip~\cite{prajwal2020lip} and IP-LAP~\cite{zhong2023identity} rely on the ground-truth upper-half region to achieve pose alignment. To enable a more equitable comparison, we conducted experiments under a fixed pose setting, and as shown in Table \ref{tab:sync-method}, our method attains the highest image quality.

\begin{table}[h]
\fontsize{10}{15}\selectfont
\centering
\caption{
Audio-driven results on HDTF~\cite{zhang2021flow} with
 fixed pose.
}
\label{tab:sync-method}
\resizebox{0.7\columnwidth}{!}{%
\begin{tabular}{ccccc}
\Xhline{1pt}
Method  & PSNR↑          & SSIM↑          & FID↓           & LMD↓          \\ \hline
Wav2Lip~\cite{prajwal2020lip} & 30.02          & 0.664          & 30.53          & 3.94          \\
IP-LAP~\cite{zhong2023identity}  & 29.47          & 0.631          & 36.14          & 3.87          \\
Ours    & \textbf{30.25} & \textbf{0.686} & \textbf{23.09} & \textbf{3.82} \\ \Xhline{1pt}
\end{tabular}%
}
\end{table}

\subsection{Robustness of the Deformation Module}
\begin{figure}[h]
\setlength{\abovecaptionskip}{0.cm}
    \centering
    \includegraphics[width=1\linewidth]{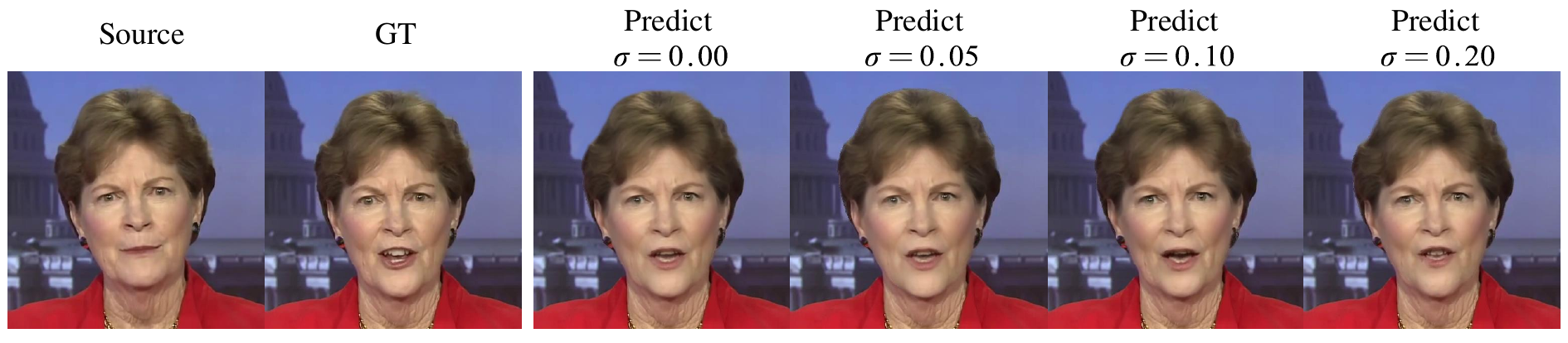}
    \caption{Visualization of adding noise to expression features.}
    \label{fig:deform_nosie}
    \vspace{-4mm}
\end{figure}

To evaluate the robustness of our Deformation module under inaccuracies in expression basis estimation, we introduce Gaussian noise into the expression basis and monitor the performance of the module. As shown in Figure \ref{fig:deform_nosie} and Table \ref{tab:noise_deform}, when the standard deviation of the Gaussian noise increases from 0 to 0.2, the predicted results remain relatively stable.

\begin{table}[h]
\fontsize{10}{15}\selectfont
\centering
\caption{
Self-reenactment results on HDTF~\cite{zhang2021flow} with varying noise intensities added to the estimated expression features.
}
\label{tab:noise_deform}
\resizebox{0.4\columnwidth}{!}{%
\begin{tabular}{ccc}
\Xhline{1pt}
Noise std & FID↓  & AED↓  \\ \hline
0.00      & 18.95 & 0.102 \\
0.05      & 19.13 & 0.104 \\
0.10      & 19.46 & 0.117 \\
0.20      & 19.74 & 0.121 \\ \Xhline{1pt}
\end{tabular}%
}
\end{table}
 \vspace{-5mm}

% %

% %

%

%

%
%

% %

% %
% %